\documentclass[letterpaper]{article} 
\usepackage{aaai24}  
\usepackage{times}  
\usepackage{helvet}  
\usepackage{courier}  
\usepackage[hyphens]{url}  
\usepackage{graphicx} 
\usepackage{natbib}  
\usepackage{caption} 
\usepackage{algorithm}
\usepackage{algorithmic}

\usepackage{newfloat}
\usepackage{listings}
\DeclareCaptionStyle{ruled}{labelfont=normalfont,labelsep=colon,strut=off} 
\floatstyle{ruled}
\newfloat{listing}{tb}{lst}{}
\floatname{listing}{Listing}
\nocopyright
\usepackage{amsfonts}
\usepackage{amsmath}
\usepackage{algorithm}
\usepackage{algorithmic}
\usepackage{graphicx}
\usepackage{subcaption}
\usepackage{multirow}
\usepackage{array}
\usepackage{color, colortbl}
\usepackage{booktabs}

\definecolor{ashgray}{rgb}{0.7,0.75,0.71}
\definecolor{babypink}{rgb}{0.96,0.76,0.76}
\definecolor{taupegray}{rgb}{0.55, 0.52, 0.54}
\definecolor{gainsboro}{rgb}{0.86, 0.86, 0.86}
\definecolor{lightgray}{rgb}{0.83, 0.83, 0.83}
\definecolor{indigo(dye)}{rgb}{0.0, 0.25, 0.42}
\definecolor{silver}{rgb}{0.75, 0.75, 0.75}

\title{Logical Specifications-guided Dynamic Task Sampling\\ for Reinforcement Learning Agents}
\author{
Yash Shukla$^{1}$ \hspace{3em} Tanushree Burman$^{1}$ \hspace{3em} Abhishek N. Kulkarni$^{2}$ \\ \hspace{-2em} Robert Wright$^{3}$ \hspace{2.5em} Alvaro Velasquez$^{4}$ \hspace{4em} Jivko Sinapov$^{1}$
}
\affiliations{
    \textsuperscript{\rm 1}Tufts University \hspace{1em}
    \textsuperscript{\rm 2}University of Texas at Austin \hspace{1em}
    \textsuperscript{\rm 3}Georgia Tech Research Institute \hspace{1em}
    \textsuperscript{\rm 4}University of Colorado Boulder
    
}

\usepackage{bibentry}

\begin{document}

\maketitle
\begin{abstract}
Reinforcement Learning (RL) has made significant strides in enabling artificial agents to learn diverse behaviors. However, learning an effective policy often requires a large number of environment interactions. To mitigate sample complexity issues, recent approaches have used high-level task specifications, such as Linear Temporal Logic (LTL$_f$) formulas or Reward Machines (RM), to guide the learning progress of the agent. In this work, we propose a novel approach, called Logical Specifications-guided Dynamic Task Sampling (LSTS), that learns a set of RL policies to guide an agent from an initial state to a goal state based on a high-level task specification, while minimizing the number of environmental interactions. Unlike previous work, LSTS does not assume information about the environment dynamics or the Reward Machine, and dynamically samples promising tasks that lead to successful goal policies. We evaluate LSTS on a gridworld and show that it achieves improved time-to-threshold performance on complex sequential decision-making problems compared to state-of-the-art RM and Automaton-guided RL baselines, such as Q-Learning for Reward Machines and Compositional RL from logical Specifications (DIRL). Moreover, we demonstrate that our method outperforms RM and Automaton-guided RL baselines in terms of sample-efficiency, both in a partially observable robotic task and in a continuous control robotic manipulation task.
\end{abstract}

\section{Introduction}

Agents are now capable of learning optimal control behavior for a broad spectrum of tasks, ranging from Atari games to robotic manipulation tasks, thanks to recent advancements in Reinforcement Learning (RL). Despite the progress made in RL, learning an optimal task policy using model-free RL techniques still suffers from sample complexity issues because of sparse reward settings and unknown transition dynamics~\citep{lattimore2013sample}. These challenges further intensify in long-horizon settings, where the agent needs to perform a series of correct sequential decisions to achieve the goal. Certain tasks (such as - robot needs to make dinner only if it bought groceries in the afternoon) require the agent to \textit{encode} and \textit{remember} its episodic history (whether the groceries were bought) in order to solve the task effectively. 
To alleviate this issue in complicated tasks, several lines of work have explored representing the goal using an intricately shaped reward function that guides the agent toward the goal~\citep{grzes2017reward}. However, generating such a reward function requires the human engineer to assign `importance' weights to certain aspects of the task, which is time consuming and assumes knowledge on which aspects of the task are important. Poorly engineered reward functions can lead to local optima, where the agent learns to satisfy only a subset of goals and ignores the rest. 

Recent research has investigated representing the goal using high-level specification languages, such as finite-trace Linear Temporal Logic (LTL$_f$)~\citep{de2013linear}, Reward Machines (RM)~\citep{icarte2022reward}, SPECTRL~\cite{spectrl} that allow defining the goal of the task using a graphical representation of sub-tasks. The high-level objective is known before commencing the task, and the graphical representation allows the agent to learn policies that achieve easier sub-goals initially, and build upon them to achieve complex goals. Encoding the task using a graphical structure allows us to tackle the problem in a Markovian manner by tracking the history as a part of the state space~\cite{afzal2023ltl}, thereby allowing the agent to keep track of its episodic history. For instance, if the task for a robot is to reach kitchen and then make dinner, the graphical structure of the task obtained from the high-level specification allows the agent to reason whether it has reached the kitchen before it can commence its policy for making dinner. RM approaches still require human guidance in defining the reward structure, which is dependent on knowing how much reward should be assigned for accomplishing each sub-goal. The process of designing the reward structure assumes that the human engineer is aware of how much should reward should the agent receive when it accomplishes the sub-goals in particular order. This assumption is infeasible in scenarios when the structure of the environment or the exact order in which the sub-goals must be achieved is unknown in advance. In contrast, our method does not require access to the reward structure.

\begin{figure*}[t]
	\centering
	\begin{minipage}{.25\textwidth}
		\centering
		\includegraphics[width=\textwidth,height=0.9\textwidth]{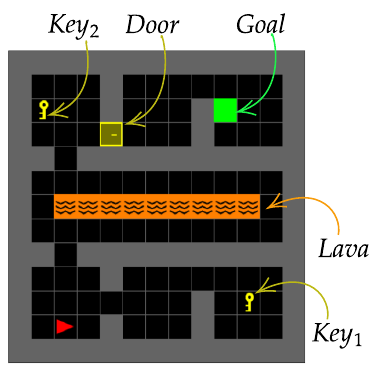}
        \subcaption{Gridworld domain}
        \label{fig:minigrid}
	\end{minipage}%
	\centering
	\begin{minipage}{.33\textwidth}
		\centering
		\includegraphics[width=\textwidth,height=0.55\textwidth]{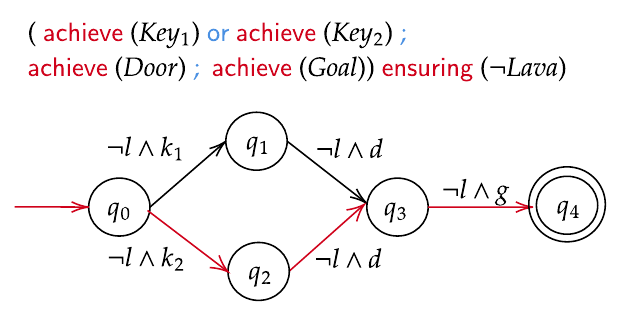}
        \subcaption{\textsc{SPECTRL} formula and its corresponding DAG. The DAG excludes all self-loops and transitions to a sink state.}
        \label{fig:minigrid-ltl}
	\end{minipage}%
	\hfill
	\centering
	\begin{minipage}{.31\textwidth}
		\centering
		\includegraphics[width=\textwidth,height=0.7\textwidth]{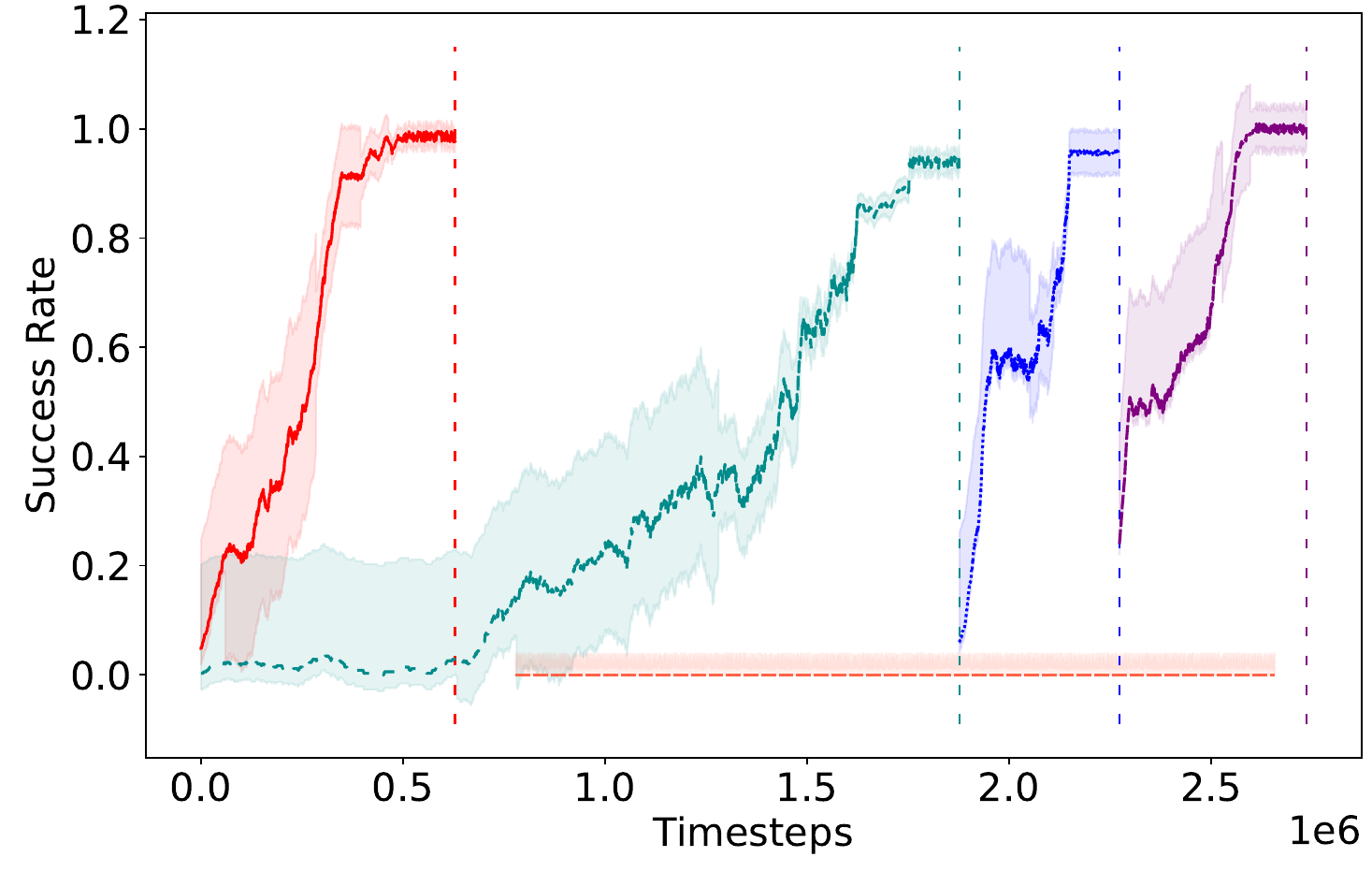}
        \subcaption{Learning curve}
        \label{fig:minigrid-learning}
	\end{minipage}%
 	\centering
	\begin{minipage}{.11\textwidth}
		\centering
		\includegraphics[width=1\textwidth,height=1\textwidth]{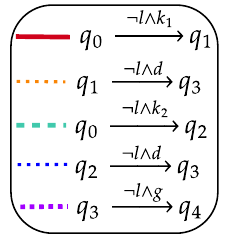}
	\end{minipage}%
	\caption{(a) Task environment. The agent (denoted using red triangle) has to \textit{pick-up} one of the keys and \textit{toggle} the door open to reach the goal state; (b) The \textsc{SPECTRL} formula for the task and its DAG. Formulas $l$, $k_1$, $k_2$, $d$ and $g$ correspond to $Lava$, $Key_1$, $Key_2$, $Door$ and $Goal$ respectively; (c) Learning curves for individual sub-tasks (averaged over 10 trials) generated using \emph{LSTS}. The path chosen by \emph{LSTS} is highlighted in red in Fig.1(b)} 
 \label{fig:overview}
\end{figure*}

Another method, Compositional RL from Logical Specifications (\textsc{DiRL})~\citep{jothimurugan2021compositional} mitigates the reward assignment issue by using Dijkstra's algorithm to determine which sub-tasks (edges) should be explored in the SPECTRL DAG graph~\cite{spectrl} in order to learn policies to reach nodes in the DAG that yield the highest success rate.~\textsc{DiRL} requires the agent to learn RL policies for satisfying \textit{all} outgoing edge propositions (each edge encodes a sub-task) from such nodes. However, this approach requires the agent to explore a sub-task for a manually specified number of interactions, which requires knowledge about the task complexity.~\textsc{DiRL} can be sample-inefficient as it attempts to learn unproductive policies on unpromising some sub-tasks, as the agent has to spend the defined number of interactions learning a policy for the sub-task. Unlike ~\textsc{DiRL}, our approach is sample-efficient as it finds unpromising sub-tasks based on the learning progress of the sub-tasks, and discards them; saving costly interactions and converging to a successful policy faster. This problem of minimizing the overall number of interactions while learning a set of successful policies is non-trivial as the problem equates to finding the shortest path in a graph where the true edge weights are unknown \textit{a priori}~\citep{szepesvari2004shortest}. In our case, the edge weight denotes the total number of environmental interactions required by the agent to learn a successful policy for the sub-task encoded by the edge, in which the agent must induce a visit to a state where certain properties hold true.
And, we can sample interactions for a sub-task only if we have a policy to reach the edge's source node from the start node of the graph, making the learning process more sample-inefficient. 


To address the above challenges, we present Logical Specifications-guided Dynamic Task Sampling (\emph{LSTS}). We begin with a high-level objective represented using ~\textsc{SPECTRL} specification formulas which can equivalently be represented using directed acyclic graphs (DAG)~\cite{spectrl}. The DAG structure encodes memory, helping the agent understand what events of interest have occurred in the past, and which events must occur to reach the accepting states. Our key insight is to learn RL policies for sub-tasks defined using cost-to-goal formmulation of the edges of the DAG. Specifically, the agent transitions from the node $q$ to $p$ in the DAG when the propositional logic formula labeling the edge $(q, p)$ evaluates to true. We use the set of propositional logic formulas labeling the outgoing edges from a given node in DAG to define sub-tasks. The trajectory induced by a successful RL policy for the sub-task enables the agent's high-level state in the DAG to transition from the source node to the destination node of the edge defining the sub-task. We utilize an adaptive Teacher-Student learning approach, where, (1) the Teacher agent uses its high-level policy to actively sample a sub-task for the Student agent to explore. The high-level policy considers the DAG representation and the Student agent's expected performance on all the sub-tasks. The goal of the Teacher agent is to satisfy the high-level objective in the fewest number of environmental interactions, and (2) The Student agent explores the environment for few interactions (much fewer than the interactions required to learn a successful policy for the sub-task) while simultaneously updating its low-level RL policy for the sampled sub-task. Next, the Teacher observes the Student's performance and updates its high-level policy. Steps (1) and (2) continue alternately until the Student agent learns a set of successful policies to satisfy the high-level objective.

\textbf{Running example:} As an example, let us look at the environment shown in Fig.~\ref{fig:minigrid}. The goal for the agent is to collect \emph{any} of the two \emph{Keys}, followed by opening the \emph{Door} and then reaching the \emph{Goal} while avoiding the \emph{Lava} at all times. The task's high-level objective ($\phi$) is represented using the \textsc{SPECTRL} formula and its corresponding DAG representation $\mathcal{G}_{\phi}$ in Fig.~\ref{fig:minigrid-ltl}. The DAG does not contain information about the environment configuration, such as: the optimal number of interactions required to reach $Door$ from $Key_1$ are much higher compared to the interactions required to reach $Door$ from $Key_2$, making the $Key_1$ to $Door$ trajectory sub-optimal. Hence, it is crucial to prevent any additional interactions the agent spends in learning a policy for the sub-task defined by the edge $q_1 \xrightarrow{\neg l \wedge d} q_3$ as the path $q_0 \xrightarrow{\neg l \wedge k_1} q_1 \xrightarrow{\neg l \wedge d} q_3 \xrightarrow{\neg l \wedge g} q_4$ will always be sub-optimal. In our proposed approach \emph{LSTS}, the Student agent begins with the \textit{aim} of learning two distinct RL policies: $\pi_1$ for the task of visiting $Key_1$ and $\pi_2$ for the task of visiting $Key_2$, both avoiding $Lava$. The Teacher agent initially samples evenly from these two sub-tasks for the Student but later biases its sampling toward the sub-task on which the Student agent shows higher learning potential. Once the Student agent learns a successful policy for one of the sub-tasks (let's say the learned policy $\pi_1^*$ corresponding to the sub-task defined by the transition $ q_0 \xrightarrow{\neg l \wedge k_1} q_1$), the Teacher does not sample that task anymore, identifies the next task(s) for the Student using the DAG representation, and appends them to the set of tasks it is currently sampling (in this case, the only next task is: $q_1 \xrightarrow{\neg l \wedge d} q_3$). Since the Student agent only has access to the state distribution over $q_0$, it follows the trajectory given by $\pi_1^*$ to reach a state that lies in the set of states where the proposition $\neg Lava \wedge Key_1$ holds true before commencing its learning for the policy ($\pi_3$) for $q_1 \xrightarrow{\neg l \wedge d} q_3$. If the Student agent learns the policies $\pi_2^*$ for satisfying the sub-task defined by $ q_0 \xrightarrow{\neg l \wedge k_2} q_2$ and $\pi_4^*$ for $ q_2 \xrightarrow{\neg l \wedge d} q_3$ before learning $\pi_3$, it effectively has a set of policies to reach the node $q_3$. Thus, the Teacher will now only sample the next task for the Student in the DAG representation $q_3 \xrightarrow{\neg l \wedge g} q_4$, as learning RL policies for paths that reach $q_3$ are effectively redundant. This process continues iteratively until the Student agent learns a set of policies that reach the goal node ($q_4$) from the start node ($q_0$). The learning curves in Fig.~\ref{fig:minigrid-learning} empirically validate the running example. As evident from the learning curves, the Student agent learns policies for the path $q_0 \xrightarrow{\neg l \wedge k_2} q_2 \xrightarrow{\neg l \wedge d} q_3 \xrightarrow{\neg l \wedge g} q_4$ that produce trajectories to reach the goal node $q_4$ from the initial node $q_0$, without excessively wasting interactions on the unpromising sub-task $q_1 \xrightarrow{\neg l \wedge d} q_3$. The dashed lines in Fig.~\ref{fig:minigrid-learning} signify the interactions at which a task policy converged.

The dynamic task sampling strategy promotes \emph{LSTS} to achieve sample-efficient learning on complex tasks by identifying unpromising tasks and discarding them, saving costly interactions. Our empirical results show that \emph{LSTS} reduces environmental interactions by orders of magnitude compared to Specifications-Guided RL Baseline \textsc{DiRL}, Reward Machine-based baselines QRM~\citep{icarte2018using}, GSRS~\citep{camacho2018non}, and curriculum learning baseline TSCL~\citep{DBLP:journals/tnn/MatiisenOCS20}. We also evaluate \emph{LSTS}$^{ct}$, a modified algorithm that further improves sample efficiency by continuing exploration on a new sub-task once the sub-task goal state is reached. We perform evaluation on two simulated robotic tasks and demonstrate that \emph{LSTS} reduces the number of interactions by orders-of-magnitude when compared to automaton-guided RL baselines.    

\section{Related Work}

\textbf{Automaton-guided RL approaches} utilize temporal logic-based language specifications to define tasks~\citep{toro2018teaching,alur2022framework, xu2019transfer}. Separating policies for task sub-goals aids in abstracting knowledge that can be utilized in novel tasks~\citep{icarte2018using}, without reliance on a dense reward function. Another technique is to shape the reward in proportion to the distance from the accepting node in the automaton~\citep{camacho2018non}; however, this often leads to suboptimal reward settings. Augmenting the reward function with Monte Carlo Tree Search helps mitigate this issue~\citep{velasquez2021dynamic}. This approach requires the ability to plan-ahead in the environment, which is not always feasible. Automaton-guided RL has been used to aid navigational exploration for robotic domains~\citep{belta} and for multi-agent settings~\citep{hammond2021multi}. Generating a curriculum given the high-level objective~\citep{shukla2023agcl} requires access to the Object-Oriented MDP~\citep{diuk2008object}, which cannot be obtained if environment details are not known in advance.
\textsc{DiRL} interleaves high-level planning with RL to learn a policy for each edge, which overcomes challenges arising from poor representations~\citep{jothimurugan2021compositional}. This approach becomes inefficient in terms of number of interactions, as it requires the agent to act for a predetermined number of interactions, even if learning the task does not show any promise. Unlike previous works, in this paper, we propose an logical specifications-guided dynamic task sampling approach that does not require access to the environment dynamics or the Reward Machine, and efficiently samples tasks that show promise toward the high-level objective, saving interactions on unpromising tasks. \\
\textbf{Teacher-Student algorithms}~\citep{DBLP:journals/tnn/MatiisenOCS20} have been previously studied in Curriculum Learning literature~\citep{narvekar2020curriculum,shukla2022acute, shukla2024lgts} and in the Intrinsic Motivation literature~\citep{oudeyer2009intrinsic}.  These approaches utilizes a bi-level structure, where a Teacher agent suggests tasks to the Student based on the Student's demonstrated potential. This approach facilitates the learning of simpler tasks initially, enabling the transfer of knowledge to more complex tasks and reducing the number of interactions required to attain a successful policy. However, these methods typically focus on optimizing a curriculum for learning a single policy, which may not be suitable for tasks with extended temporal aspects. In contrast, we propose a Logical Specifications-driven Teacher-Student learning approach that targets the learning of policies for promising automaton transitions, thereby enhancing sample efficiency compared to conventional approaches.

\section{Theoretical Framework}


\textbf{Episodic MDP.} An episodic labeled Markov Decision Process (MDP) $M$ is a tuple $(\mathcal{S}, \mathcal{A}, P, R, \mathcal{S}_0, \gamma, K, \mathcal{P}, L)$, where $\mathcal{S}$ is the set of states, $\mathcal{A}$ is the set of actions, $P(s'|s,a)$ denotes the transition probability of reaching state $s' \in \mathcal{S}$ from $s \in \mathcal{S}$ using action $a \in \mathcal{A}$, $R: \mathcal{S} \times A \times \mathcal{S} \rightarrow \mathbb{R}$ is the reward function, $\mathcal{S}_0$ is the initial state distribution, $\gamma \in  [0,1]$ is the discount factor, $K$ is the maximum number of interactions in any episode, $\mathcal{P}$ is a set of predicates, and $L: \mathcal{S} \rightarrow 2^{\mathcal{P}}$ is a labeling function that maps a state $s \in \mathcal{S}$ to a subset of predicates that are true in that state. During each interaction, the agent observes the current state $s$ and selects an action $a$ according to its policy function $\pi(a|s, \theta)$ with parameters $\theta$. The agent transitionsn to a new state $s' \in \mathcal{S}$ with probability $P(s' \mid s, a)$. The agent's goal is to learn an \emph{optimal policy $\pi^*$} that maximizes the discounted return $G_0 = \sum^{K}_{k = 0}\!\gamma^k\! R(s'_k,a_k,s_k) $ until the end of the episode, which occurs after at-most $K$ interactions.\\
\\
\textbf{High level specification language}:
In our framework, we adopt the specification language~\textsc{SPECTRL} to articulate reinforcement learning tasks~\cite{spectrl}. A specification $\phi$ in ~\textsc{SPECTRL} is a logical formula applied to trajectories, determining whether a given trajectory $\zeta = (s_0, s_1 , \ldots)$ successfully accomplishes a desired task. Mathematically, $\phi$ can be depicted as a function $\phi:\mathcal{Z}\to\mathbb{B}$, where $\mathbb{B}=\{\textsc{True},\textsc{False}\}$ and $\mathcal{Z}$ is the set of all trajectories.

Formally, a specification is defined over a set of \emph{atomic predicates} $\mathcal{P}_0$. Each $p \in \mathcal{P}_0$ is associated with a function $f_p:S\to\mathbb{B}$. The agent's MDP state $s$ satisfies $p$ (denoted by $s\models p$) when $f_p(s)=\texttt{True}$ (in other words, $p \subseteq L(s)$). 


The set of \emph{predicates} $\mathcal{P}$ comprises conjunctions and disjunctions of atomic predicates $\mathcal{P}_0$. A predicate $b\in\mathcal{P}$ follows the grammar $b ::= p \mid (b_1 \wedge b_2) \mid (b_1 \vee b_2)$, where $p\in\mathcal{P}_0$. Each predicate $b\in\mathcal{P}$ corresponds to a function $f_b:S\to\mathbb{B}$ defined naturally over Boolean logic.

The syntax of \textsc{SPECTRL} specifications is given by
\[ \phi ::= \texttt{achieve} \:\: b \mid \phi_1 \:\: \texttt{ensuring} \:\: b \mid \phi_1; \phi_2 \mid \phi_1 \:\: \texttt{or} \:\: {\phi_2}, \]
where $b\in\mathcal{P}$. Here, $\texttt{achieve}$ and $\texttt{ensuring}$ correspond to `eventually' and `always' operators in temporal logic. Each specification $\phi$ corresponds to a function $f_{\phi}:\mathcal{Z}\to\mathbb{B}$, and $\zeta\in\mathcal{Z}$ satisfies $\phi$ (denoted $\zeta\models\phi$) if $f_{\phi}(\zeta) := \textsc{True}$. The \textsc{SPECTRL} semantics for a finite trajectory $\zeta$ of length $t$ are:
\begin{align}
    &\zeta\models \texttt{achieve} \:\: {b} \: \: \: \: \text{if} \: \: \: \exists\ i \leq t,~s_i\models b \: ( \text{or} \:\:b \subseteq L(s_i))\\
&\zeta\models \phi \:\: \texttt{ensuring} \:\: {b} \: \: \: \: \text{if} \: \: \: \exists\ i \leq t,~s_i\models b  \\ 
&\zeta\models\phi_1; \phi_2 \:\:\: \text{if} \:\:\: \exists\ i < t, ~\zeta_{0:i}\models \phi_1 \:\:\:\text{and}\:\:\: \zeta_{i+1:t}\models\phi_2 \\ 
&\zeta\models \phi_1 \:\: \texttt{or} \:\: \phi_2 \:\:\: \text{if}  \:\:\: \zeta\models\phi_1 \:\:\: \text{or} \:\:\: \zeta\models\phi_2
\end{align}

Intuitively, the condition $(1)$ signifies that the trajectory should \textit{eventually} reach a state where the predicates $b$ hold true. The condition $(2)$ signifies that the trajectory should satisfy specification $\phi$ while \textit{always} remaining in states where $b$ holds true. The condition $(3)$ signifies that the trajectory should sequentially satisfy $\phi_1$ and then $\phi_2$. The condition $(4)$ signifies that the trajectory should satisfy either $\phi_1$ or $\phi_2$. A trajectory $\zeta$ satisfies $\phi$ if there is a $t$ such that the prefix $\zeta_{0:t}$ satisfies $\phi$. 

Furthermore, each \textsc{SPECTRL} specification $\phi$ is \textit{guaranteed} to have an equivalent directed acyclic graph (DAG), called an abstract graph. An {\em abstract graph} $\mathcal{G} = (Q,E,q_0,F,\beta,\mathcal{Z}_{safe}, \kappa)$ is a directed acyclic graph (DAG) with nodes $Q$,
(directed) edges $E\subseteq Q\times Q$, initial node $q_0\in Q$, final nodes $F\subseteq Q$, subgoal region map $\beta:Q\to2^\mathcal{S}$ such that for each $q\in Q$, $\beta(q)$ is a subgoal region and \emph{safe trajectories}
$\mathcal{Z}_{safe} = \bigcup_{e \in E}\mathcal{Z}_{safe}^e$
where $\mathcal{Z}_{safe}^e\subseteq\mathcal{Z}_f$ denotes the safe trajectories for edge $e \in E$.
Intuitively, $(Q,E)$ is a standard DAG, and $q_0$ and $F$ define a graph reachability problem for $(Q,E)$. Furthermore, $\beta$ and $\mathcal{Z}_{safe}$ connect $(Q,E)$ back to the original MDP $M$; in particular, for an edge $e=q\to q'$, $\mathcal{Z}_{safe}^e$ is the set of trajectories in the MDP $M$ that can be used to transition from $\beta(q)$ to $\beta(q')$\footnote{See \textsc{DiRL}~\cite{jothimurugan2021compositional} for more details}. The function $\kappa$ labels each edge $e = q \rightarrow q'$ with the predicates $b_e$ (labeled edge denoted as $e :=q \xrightarrow[]{b_e} q'$). The agent transitions from $q$ to $q'$ when the states $s_i, s_{i+j}$ in the agent's trajectory $\zeta$ satisfy $s_i \subseteq \beta(q)$ and $b_e \subseteq L(s_{i+j})$ and $j \geq 0$.

Given a \textsc{SPECTRL} specification $\phi$, we can construct an abstract graph $\mathcal{G}_\phi$ such that, for every trajectory $\zeta \in \mathcal{Z}$, we have $\zeta\models\phi$ if and only if $\zeta\models \mathcal{G}_\phi$. Thus, we can solve the reinforcement learning problem for $\phi$ by solving the reachability problem for $\mathcal{G}_\phi$. As described below, we leverage the structure of $\mathcal{G}_\phi$ in conjunction with reinforcement learning to do so.
In summary, \textsc{SPECTRL} specifications provide a powerful and expressive means to define and evaluate reinforcement learning tasks. It allows users to specify complex conditions and requirements for successful task completion, enabling a nuanced approach to learning from specifications. \\
\\
\textbf{Problem Formulation.} Given an MDP $M$ with unknown transition dynamics and a \textsc{SPECTRL} formula $\phi$ representing the high-level task objective, let $\mathcal{G}_{\phi}$ be the DAG representing the language of $\phi$. Let $\mathsf{Paths}(q, X)$ be the set of all paths in the DAG originating in $q$ and terminating at a node in $X \subseteq Q$. The aim of this work is to learn a set of policies $\pi_{i}^*$, $i=0, \ldots, n-1$, with the following three properties: (i) Following $\pi_0^*$ results in a trajectory in the MDP that induces a transition from $q_0$ to some state $q_1 \in Q$ in the DAG, following $\pi_1^*$ results in a trajectory in MDP that induces a transition from $q_1$ to some state $q_2 \in Q$ in the DAG, and so on. (ii) The resulting path $q_0 q_1 \ldots q_n$ in the DAG terminates at a final node, \textit{i.e.}, $q_n \in F$, with probability greater than a given threshold, $\eta \in (0, 1)$. (iii) The total number of environmental interactions spent in exploring and learning sub-task policies are minimized.

\section{Methodology}~\label{sec:methodology}\\
\textbf{Sub-task definiton:}
Given the DAG $\mathcal{G}_{\phi}$ representing the language of $\phi$, we define a set of sub-tasks based on the edges of the DAG. Intuitively, given any MDP state $s \in \mathcal{S}$ and a DAG node $q \in Q$, a sub-task defined by an edge from node $q$ to $p \in Q$ defines a reach-avoid objective for the agent represented by the \textsc{SPECTRL} formula,
\[\small{\mathsf{Task}(q, p)\!\! := \!\!\texttt{achieve}(b_{(q,p)})  \: \texttt{ensuring}\left(\bigwedge\limits_{r \in \mathsf{Sc}(q), r \neq p} \!\!\!\neg b_{(q, r)}\!\!\right)} \]
where $b_{(q,p)}$ is the propositional formula labeling the edge from $q$ to $p$ in the DAG and $\mathsf{Sc}(q)$ is the set of successors of node $q$ in DAG. For example, in Fig.~\ref{fig:minigrid-ltl}, the propositional formula labeling the edge from $q_0$ to $q_1$ is $b_{(q_0, q_1)} = \neg l \land k_1$. When $e = (q, p)$, we use $\mathsf{Task}(e)$ instead of $\mathsf{Task}(q, p)$ and $b_e$ instead of $b_{(q,p)}$ for notational convenience.

Each sub-task $\mathsf{Task}(q, p)$ defines a problem to learn a policy $\pi_{(q, p)}^*$ such that, given any MDP state $s_0 \in \mathcal{S}$, following $\pi_{(q, p)}^*$ results in a trajectory $s_0 s_1 \ldots s_n$ in MDP that induces the path $q q \ldots q p$ in the DAG. That is, the agent's high-level DAG state remains at $q$ until it transitions to $p$. While constructing the set of sub-tasks, we omit transitions that lead to a `sink' state (from which final states are unreachable). 

Given the MDP $M$ with unknown transition dynamics and the \textsc{SPECTRL} objective, $\phi$, we first translate $\phi$ to its corresponding directed acyclic graphical (DAG) representation $\mathcal{G}_{\phi} = (Q,E,q_0,F,\beta,\mathcal{Z}_{safe}, \kappa)$. Next, we define the set of sub-tasks. For this, we consider the edges that lie on some path in the DAG from $q_0$ to some node in $F$. This is because any path that does not visit $F$ leads to a sink state from which the objective cannot be satisfied. Such edges are identified using breadth-first-search \cite{moore1959shortest}.
\\
\\
\textbf{LSTS Initialization:} The algorithm for \emph{LSTS} is described in Algo.~\ref{alg:LSTS}. We begin by initializing the following (lines 2-4):
(1) Set of: Active Tasks {\tt AT}, Learned Tasks {\tt LT}, Discarded Tasks {\tt DT};
(2) A Dictionary $\Pi: \mathsf{Task}(e) \rightarrow \pi_e$ that maps the sub-task $\mathsf{Task}(e)$ corresponding to edge $e$ of DAG $\mathcal{G}_{\phi}$ to a RL policy $\pi_e$; (3) A Dictionary $Q: \mathsf{Task}(e) \rightarrow \mathbb{R}$ representing the Q-Values for the Teacher agent by mapping $\mathsf{Task}(e)$ to a numerical q-value associated with $\mathsf{Task}(e)$.  

Initially, we transform $\mathcal{G}_{\phi}$ into an Adjacency Matrix denoted as $\mathcal{X}$ (line 6), then identify the tasks corresponding to all outgoing edges $\overline{E}_{q_0} \subseteq E$ originating from the initial node $q_0$ (line 7). Satisfying the predicates $b_{(q_0, q_1)} \in \kappa(\overline{E}_{q_0})$ signifies a reachability sub-task $M'$. In this sub-task, every goal state $s \in \mathcal{S}_f^{M'}$ of $M'$ satisfies the condition $b_{(q_0, q_1)} \subseteq L(s)$. The Student agent is rewarded positively for fulfilling $b_{(q_0, q_1)}$ and negatively at all other time steps. The state, action space, and transition dynamics of $M'$ mirror those of $M$. To accomplish this sub-task, the Student agent must learn a Reinforcement Learning (RL) policy $\pi_{(q_0, q_1)}^*$ ensuring a visit from $q_0$ to $q_1$ with a probability exceeding a predetermined threshold ($\eta$). Additionally, the policy must prevent unintended transitions in the Directed Acyclic Graph (DAG); for example, while obtaining $Key_1$, it should not inadvertently obtain $Key_2$.\\
\\
\textbf{Teacher-Student learning:} To begin, the Teacher Q-Values for all the sub-tasks corresponding to edges in {\tt AT} (i.e., tasks corresponding to $\overline{E}_{q_0}$) are set to zero (line 8).
We formalize the Teacher's objective of selecting the most promising sub-task within a \textit{Partially Observable Markov Decision Process} framework~\citep{KAELBLING199899}. Here, the Teacher lacks access to the complete state of the Student agent but only observes its performance on a sub-task (such as success rate or average returns). The Teacher's action involves choosing a sub-task $\mathsf{Task}(e) \in$ {\tt AT} that the Student agent should prioritize for training. Within this POMDP framework, each action taken by the Teacher is associated with a Q-Value. Generally, higher Q-Values indicate tasks where the Student agent demonstrates greater success, prompting the Teacher to prioritize sampling such tasks more frequently to achieve $\phi$ (reaching a goal node) with the fewest overall interactions.

(A) The Teacher agent samples a sub-task $\mathsf{Task}(e) \in $ {\tt AT} using the $\epsilon-$greedy exploration strategy (line 10), and (B) The Student agent trains on task $\mathsf{Task}(e)$ using its corresponding RL policy $\Pi[e]$ for \textit{a few} interactions (line 11). In one Teacher timestep, the Student trains on $x$ environmental interactions. Since the aim of the Teacher is to keep switching the Student agent's learning progress to a sub-task on which it shows the highest promise, $x << $ total number of environmental interactions required by the Student agent to learn a successful RL policy for $\mathsf{Task}(e)$. (C) The Teacher observes the Student agent's average return $g_t$ on these $x$ interactions, and updates its Q-Value for $\mathsf{Task}(e)$ (line 12) using: $Q[e] \leftarrow \alpha (g_t) + (1-\alpha)Q[e]$\\
,where $\alpha$ is the Teacher learning rate, $g_{t}$ is the average Student agent return on $\mathsf{Task}(e)$ at the Teacher timestep $t$. As learning progresses, $g_t$ increases as $t$ increases, and hence we use a constant parameter $\alpha$ to tackle the non-stationarity of a moving return distribution. Several other algorithms could be used for the Teacher strategy (e.g. Thomspson Sampling). Steps (A), (B), (C) continue successively until the policy for \textit{any} $\mathsf{Task}(e)\in ${\tt AT} task converges.\\
\\
\textbf{Sub-task convergence criteria:} The Student agent's RL policy for $\mathsf{Task}(q,p)$ is considered to be converged (line 13) if a trajectory $\zeta$ produced by the policy produces the transition in the DAG with probability Pr$_{\zeta \in \mathcal{Z}}[\zeta \mbox{ satisfies} \: \mathsf{Task}(q,p)]\:\geq \:\eta$ and $\Delta(g_t,g_{t-1})\: < \:\tau$ where $\eta$ is the expected performance and $\tau$ is a small numerical value. In simple terms, a converged policy achieves an average success rate of at least $\eta$ and shows no further improvement by maintaining consistent average returns. Similar to other RM methods, we assume access to the labeling function $L$ to check if a trajectory $\zeta$ satisfies the formula $b_{(q,p)}$ by verifying if the final state $s$ of the trajectory meets the condition $b_{(q,p)} \subseteq L(s)$. It's worth noting that the sub-goal regions may overlap, meaning the same state $s$ could satisfy propositions for multiple nodes in the Directed Acyclic Graph (DAG). Once a policy for the $\mathsf{Task}(q,p)$ converges, we add $\mathsf{Task}(q,p)$ to the set of Learned Tasks {\tt LT} and remove it from the set of Active Tasks {\tt AT} (line 14).
\\
\\
\textbf{Discarding unpromising sub-tasks:}
On learning a successful policy for the $\mathsf{Task}(q,p)$ (transition $q \xrightarrow{b_{(q,p)}} p$), we find the sub-tasks to be discarded (line 15). We find the sub-tasks corresponding to edges that: (1) lie before $p$ in a path from $q_0$ to any $q \in F$, and, (2) do not lie in a path to $q \in F$ that does not contain $p$. Intuitively, if we already have a set of policies that can generate a successful trajectory to reach the node $p$, we do not need to learn policies for sub-tasks that ultimately lead \textit{only} to $p$ (e.g., in Fig.~\ref{fig:overview} if we have successful policies for $\mathsf{Task}(q_0,q_2) $ and $\mathsf{Task}(q_2,q_3)$, we can discard $\mathsf{Task}(q_0,q_1) $ and $\mathsf{Task}(q_1,q_3)$). We add all such sub-tasks to the set of Discarded Tasks {\tt DT} (line 16). As an extension, in the limit, an optimal policy can be found by not completely discarding such sub-tasks, but rather biasing away from them so that they would still be explored. \\
\begin{algorithm}[t]
\caption{ \emph{LSTS} ( $\mathcal{G}_{\phi}, M, \eta, x$ )}
\label{alg:LSTS}
\raggedright \textbf{Output}: Set of learned policies : $\Pi^*$, Edge-Policy Dictionary $\mathcal{P}$\\
\begin{algorithmic}[1] 
\STATE \textbf{Placeholder Initialization}:\\
\STATE Sets of: Active Tasks ({\tt AT}) $\leftarrow \emptyset$; \\Learned Tasks ({\tt LT}) $\leftarrow \emptyset$; Discarded Tasks ({\tt DT}) $\leftarrow \emptyset$
\\
\STATE Edge-Policy Dictionary $\Pi :  \mathsf{Task}(e) \rightarrow \pi$
\STATE Teacher Q-Value Dictionary: $Q : \mathsf{Task}(e) \rightarrow -\infty $
\STATE \textbf{Algorithm:}
\STATE $\mathcal{X} \leftarrow $ {\tt Adjacency\_Matrix} $(\mathcal{G}_{\phi})$ 
\STATE {\tt AT} $\leftarrow $ {\tt AT} $\cup$ $\{ \mathsf{Tasks}(\mathcal{X}[q_0])\}$ 
\STATE $\forall \:  \mathsf{Task}(e) \in  $ {\tt AT}   $: Q[e] = 0$  
\WHILE{True} 
\STATE $e \leftarrow $ {\tt Sample}$(Q)$ 
\STATE $\Pi[e], g \leftarrow $ {\tt Learn}$(M, \mathcal{G}_{\phi}, e, x, \mathcal{P})$ 
\STATE {\tt Update\_Teacher}$(Q, e, g)$  
\IF{{\tt Convergence}($Q, e, g, \eta$)}
\STATE $\Pi^* \leftarrow \Pi^* \cup  \Pi[e]$ ;  {\tt LT} $\leftarrow$ {\tt LT} $ \cup \{\mathsf{Task}(e)\}$ ;  \\{\tt AT} $\leftarrow$ {\tt AT} $ \setminus \{\mathsf{Task}(e)\}$
\STATE $\mathsf{Tasks}(\overline{E}_{DT}) \leftarrow $ {\tt Discarded\_Tasks}$(\mathcal{X}, e)$
\STATE {\tt DT} $\leftarrow$ {\tt DT} $\cup \:  \mathsf{Tasks}(\overline{E}_{DT})$ 
\STATE $\mathsf{Tasks}(\overline{E}_{AT}) \leftarrow$ {\tt Next\_Tasks} $(\mathcal{X}, e,$ {\tt DT}) 
\IF{$|\mathsf{Tasks}(\overline{E}_{AT})| = 0$} 
\STATE {\tt break} 
\ENDIF
\STATE $\forall\: \mathsf{Task}(\overline{e})\: \in\: \mathsf{Tasks}(\overline{E}_{AT}) : Q[\overline{e}] = 0 $  
\STATE {\tt AT}$ \leftarrow$  {\tt AT} $\cup \: \mathsf{Tasks}(\overline{E}_{AT})$ 
\ENDIF
\ENDWHILE
\STATE \textbf{return} $\Pi^*, \Pi$
\end{algorithmic}
\end{algorithm}
\\
\textbf{Traversing in the DAG until $\phi$ satisfied:} Next, we identify the subsequent set of tasks $\mathsf{Tasks}(\overline{E}_{AT})$ within the DAG to include in the {\tt AT} set (line 17). This is determined by pinpointing sub-tasks corresponding to all outgoing edges from $p$. Given that the edge $e_{q,p}$ represents the transition $q \xrightarrow{b_{(q,p)}} p$, we have learned policies that can produce a trajectory leading to a state where $b_{(q,p)}$ holds true. $\mathsf{Tasks}(\overline{E}_{AT})$ consists of $\mathcal{X}[p] \setminus ${\tt DT} (where `$\setminus$' denotes set-minus), which indicates sub-tasks corresponding to outgoing edges from $p$ not included in the {\tt DT} set.

Once $\mathsf{Tasks}(\overline{E}_{AT})$ is identified, we assign Teacher Q-values of $0$ to all $\mathsf{Task}(\overline{e})\in\mathsf{Tasks}(\overline{E}_{AT})$ to ensure the Teacher samples these tasks (line 21). In our episodic scenario, each episode initiates from a state $s\sim\mathcal{S}_0$ where the propositions for $q_0$ are true. If the current sampled sub-task is $\mathsf{Task}(p,r)$, the agent follows a trajectory utilizing learned policies from $\Pi^*$ to reach a state where the propositions for reaching DAG node $p$ are true (i.e., $s \in \beta(p)$). Subsequently, the agent endeavors to learn a separate policy for $\mathsf{Task}(p,r)$.

These aforementioned steps (lines 9-24) continue iteratively until $|\mathsf{Tasks}(\overline{E}_{AT})| = \emptyset$. This signifies that there are no further tasks to incorporate into our sampling strategy, and we have arrived at a node $q \in F$. Hence, we exit the {\tt while} loop (line 19) and return the set of learned policies $\Pi^*$ along with the edge-policy dictionary $\Pi$ (line 25). From $\Pi$ and $\Pi^*$, we obtain an ordered list of policies $\Pi^*_{list} = [\pi_{(q_1,q_2)}, \pi_{(q_2,q_3)}, \ldots, \pi_{(q_{k-1},q_k)}]$ such that sequentially following $\pi \in \Pi^*_{list}$ generates trajectories satisfying the \textsc{SPECTRL} objective $\phi$\footnote{More details: https://github.com/shukla-yash/lsts-icaps-24}.
\\
\textbf{Guarantee}: Given the ordered list of policies $\Pi^*_{list}$, we can generate a trajectory $\zeta$ in the task $M$ with  Pr$_{\zeta \in \mathcal{Z}}[\zeta \mbox{ satisfies } \phi ] \geq \eta$ (Details in Appendix B). 

\section{Experimental Results}

We aim to answer the following questions: (Q1) Does \emph{LSTS} yield sample efficient learning compared to state-of-the-art baselines?
(Q2) After reaching a sub-task goal state, can we sample a new sub-task to continue training and improve sample efficiency? 
(Q3) Does \emph{LSTS} yield sample efficient learning for complex robotic tasks with partially observable or continuous control settings? (Q4) How does \emph{LSTS} scale to complex search-and-rescue scenarios? 

\subsection{LSTS - Gridworld Domain}~\label{sec:results-gridworld}
To answer (Q1), we evaluated \emph{LSTS} on a Minigrid~\citep{minigrid} inspired domain with the \textsc{SPECTRL} objective:
\begin{multline}
\phi^{grid}_f :=( \texttt{achieve}(k_{1}) \: \texttt{or achieve} (k_{2}) ; \\ 
\texttt{achieve}(d) ; \: \texttt{achieve}(g)) \texttt{ensuring} (\neg l)
\end{multline}

\begin{table*}
    
	\begin{minipage}{0.499\linewidth}
		\centering
	\begin{tabular}{>{\centering}p{1.1cm}>{\centering}p{2.6cm}>{\centering}p{2.4cm}p{0.05cm}}
\arrayrulecolor{taupegray}\toprule
\textbf{Approach} & \textbf{$\#$ Interactions} \\ (Mean $\pm$ SD) & \textbf{Success Rate }\\ (Mean $\pm$ SD) \tabularnewline
\midrule
\rowcolor{gainsboro}
\emph{LSTS}  & $(2.72 \pm 0.31)\times10^6$ &\centering $0.96 \pm 0.02$\tabularnewline
\rowcolor{gainsboro}
\emph{LSTS}$^{ct}$  &  $(2.45 \pm 0.25)\times10^6$ &\centering $0.95 \pm 0.01$\tabularnewline
\textsc{DiRL}$^c$  & $(4.06 \pm 0.37)\times10^6$ &\centering $0.95 \pm 0.03$\tabularnewline
\textsc{DiRL} & $(5.47 \pm 0.40)\times10^6$ &\centering $0.94 \pm 0.01$ \tabularnewline
QRM  & $5\times10^7$ &\centering $0.05 \pm 0.04$ \tabularnewline
GSRS  & $5\times10^7$ &\centering $0 \pm 0$ \tabularnewline
TSCL  & $5\times10^7$ &\centering $0 \pm 0$ \tabularnewline
LFS  & $5\times10^7$ &\centering $0 \pm 0$ \tabularnewline
\bottomrule
\end{tabular}
		\caption{Table comparing $\#$interactions $\&$ success rate. \\\emph{LSTS} (highlighted) outperfomed all baselines}
		\label{table:gridworld}
	\end{minipage}
	\begin{minipage}{0.48\linewidth}
		\centering
		\includegraphics[width=0.65\textwidth,height=0.45\textwidth]{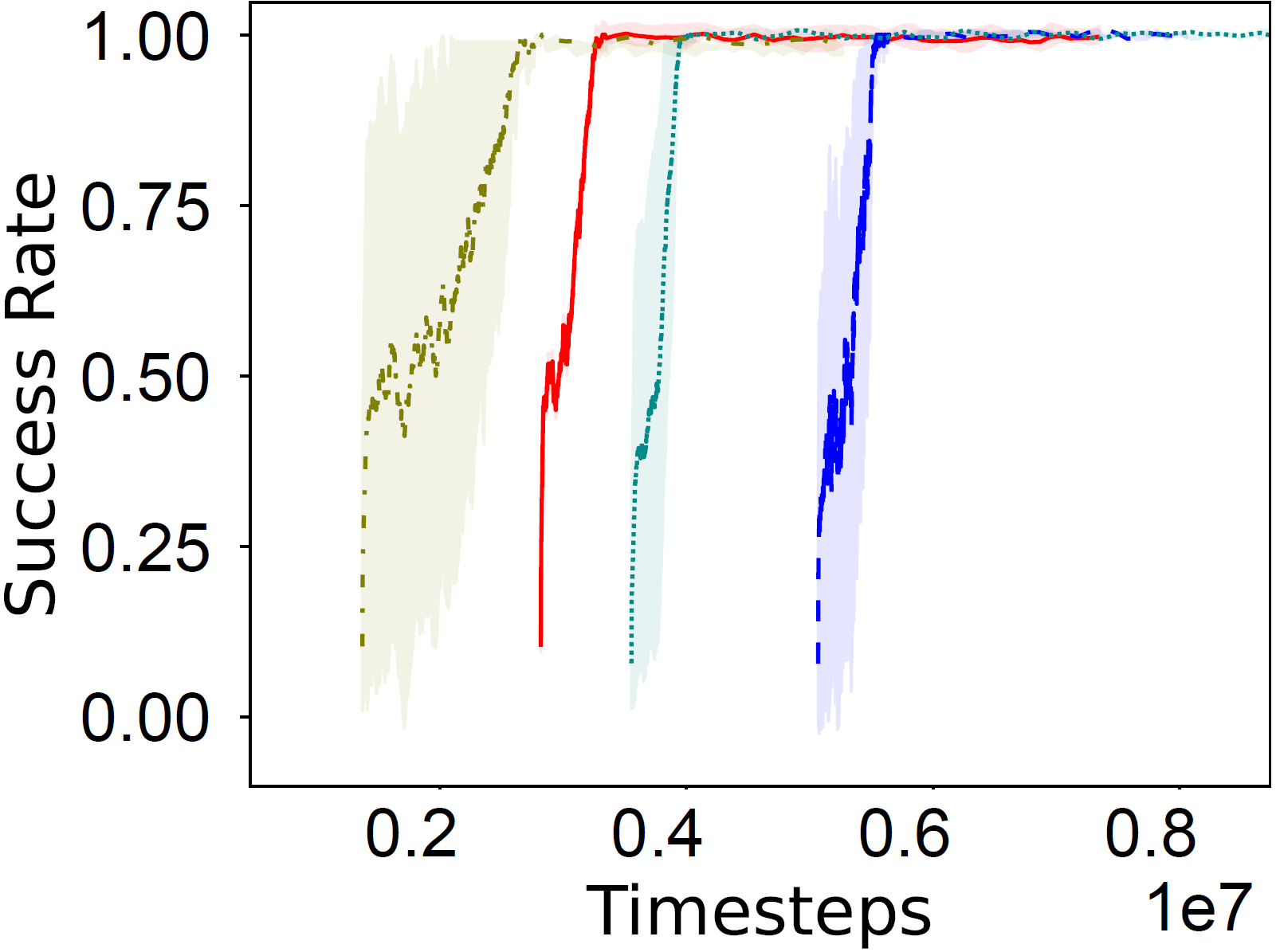}

    \includegraphics[width=0.85\textwidth,,height=0.05\textwidth]{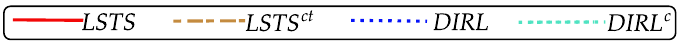}  
		\captionof{figure}{Averaged over 10 trials: Learning curves for approaches with converged policies. }
		\label{fig:lc-gridworld}
	\end{minipage}
\end{table*}

where $k_1, k_2, d, g, l$ represent the atomic propositions $Key_1, Key_2, Door, Goal, Lava$ respectively. The environment and its $\phi$ are given in Fig.~\ref{fig:overview}. The agent needs to \textit{pick-up} \textit{any} of the \emph{Keys} before heading to the \emph{Door}. After \emph{toggling} the \emph{Door} open, the agent needs to reach the \emph{Goal}. At all times, the agent needs to avoid the \emph{Lava}. For our experiments, we assume an episodic setting where an episode ends if the agent touches the \emph{Lava} object, reaches the \emph{Goal} or exhausts the number of allocated interactions.  

The problem is complex as the agent needs to perform a series of high-level correct actions to satisfy $\phi^{grid}_f$. The action space of the agent consists of three navigational actions: \emph{move forward}, \emph{rotate left} and \emph{rotate right}. The agent can also perfom: \emph{pick-up} action that adds a \emph{Key} to the agent's inventory if the agent is facing the \emph{Key}, \emph{drop} places the \emph{Key} in the next grid if grid is empty and \emph{Key} is present in the inventory, and, a \emph{toggle} action for the \emph{Door} (closed $\leftrightarrow$ open) only if the agent is holding the \emph{Key}. In this experiment, we assume a fully-observable setting where the environmental state is an image-like low-level encoding of the state. For each cell in the grid, the encoding returns an integer that describes the item occupying the grid, along with any additional information (e.g., the \emph{Door} can be open or closed).

We employ PPO~\citep{DBLP:journals/corr/SchulmanWDRK17} for the Student RL agent, which works for discrete and continuous action spaces. We consider a standard actor-critic architecture with 2 convolutional layers followed by 2 fully connected layers. For \emph{LSTS}, the reward function is sparse. The agent gets a reward of $(1 - 0.9\frac{(interactions \:taken)}{(interactions \:allocated)})$ (where $interactions\:allocated = 100$) if it achieves the goal in the sub-task, and a reward of $0$ otherwise.
\\
\textbf{Baselines:} We compare our \emph{LSTS} method with six baseline approaches: learning from scratch (LFS), Reward Machine-based (RM) baselines: GSRS~\citep{camacho2018non}, QRM~\citep{icarte2018using}; and Compositional RL from Logical Specifications (\textsc{DiRL})~\citep{jothimurugan2021compositional}.
All the baselines are implemented using the RL algorithm (PPO), described above. GSRS assigns reward inversely proportional to the distance from the RM goal node. QRM employs a separate Q-function for each node in the RM, and \textsc{DiRL} uses Dijkstra's algorithm (edge cost is the average RL policy success rate) to guide the agent in choosing a path from the specification graph.
For the fifth baseline, we modify \textsc{DiRL} such that instead of manually specifying a limit on the number of interactions, which needs to be fine-tuned to suit the task, we stop learning a sub-task once it has reached the convergence criteria defined in Section~\ref{sec:methodology}. We call this modified baseline as \textsc{DiRL}$^c$. The sixth baseline (TSCL~\citep{DBLP:journals/tnn/MatiisenOCS20}) follows a curriculum learning strategy where the Teacher samples most promising task without the use of any automaton to guide the learning progress of the agent. (More details in Appendix C)  \\
\\
The results in Table~\ref{table:gridworld} and Fig.~\ref{fig:lc-gridworld} (averaged over 10 trials) show that \emph{LSTS} reaches a successful policy quicker compared to all baselines. \emph{LSTS}$^{ct}$ is a modified version of $\emph{LSTS}$ and is described below. The learning curves in Fig.~\ref{fig:lc-gridworld} have an offset on the x-axis to account for the interactions in the initial sub-tasks before moving on to the final task in the specification, signifying strong transfer. Our custom baseline, \textsc{DiRL}$^c$ is more sample-efficient than \textsc{DiRL}, and both outperform other baselines, which do not learn a meaningful policy. We performed an unpaired t-test~\citep{kim2015t} to compare \emph{LSTS} against the best performing baselines at the end of $10^7$ training steps and we observed statistically significant results (95$\%$ confidence). Thus, \emph{LSTS} not only achieves a better success rate, but also converges faster (statistical significance result details in Appendix D). Time-to-threshold metric is defined as the difference in number of interactions between two approaches to reach a desired performance~\citep{narvekar2020curriculum}. From Fig.~\ref{fig:lc-gridworld}, we see that the time-to-threshold between \emph{LSTS} and the best-performing baseline~\textsc{DiRL}$^c$ is $1.34\times10^6$ interactions for $95\%$ success rate.

\begin{figure*}[t]
	\begin{minipage}{0.25\textwidth}
		\centering
		\includegraphics[width=0.8\textwidth,height=0.65\textwidth]{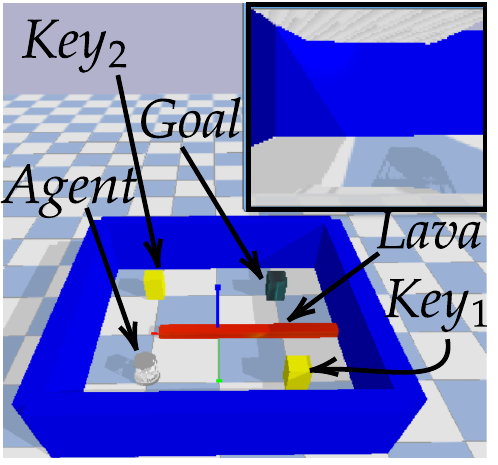}
        \subcaption{TurtleBot domain}
        \label{fig:turtlebot}
	\end{minipage}%
	\centering
	\begin{minipage}{.25\textwidth}
		\centering
		\includegraphics[width=\textwidth,height=0.65\textwidth]{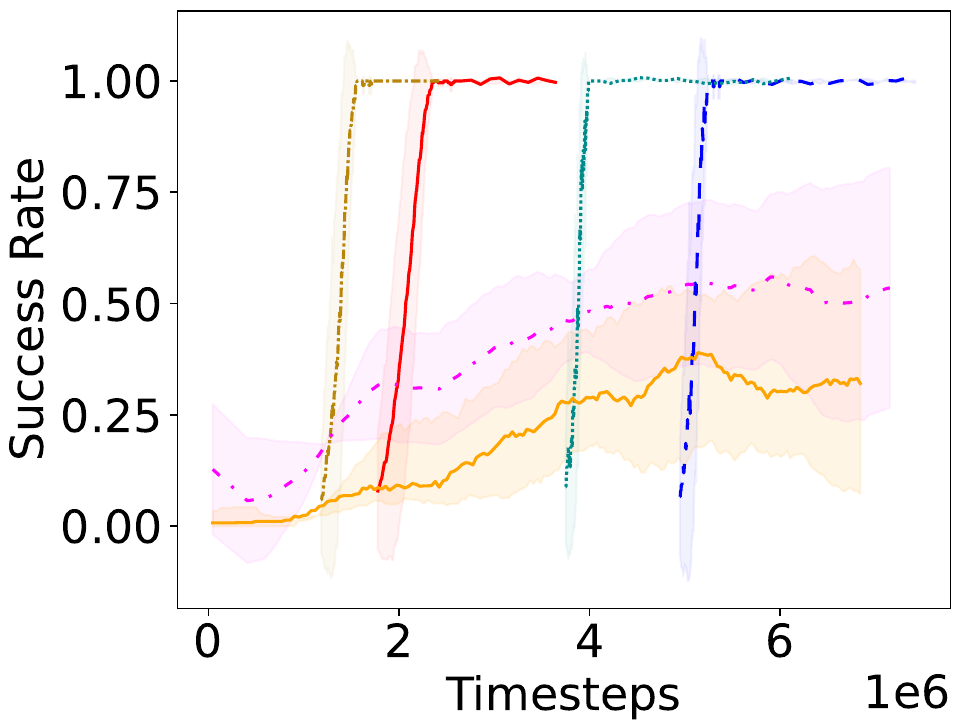}
        \subcaption{Turtlebot results}
        \label{fig:lc-turtlebot}
	\end{minipage}%
	\hfill
	\begin{minipage}{.25\textwidth}
		\centering
		\includegraphics[width=0.8\textwidth,height=0.65\textwidth]{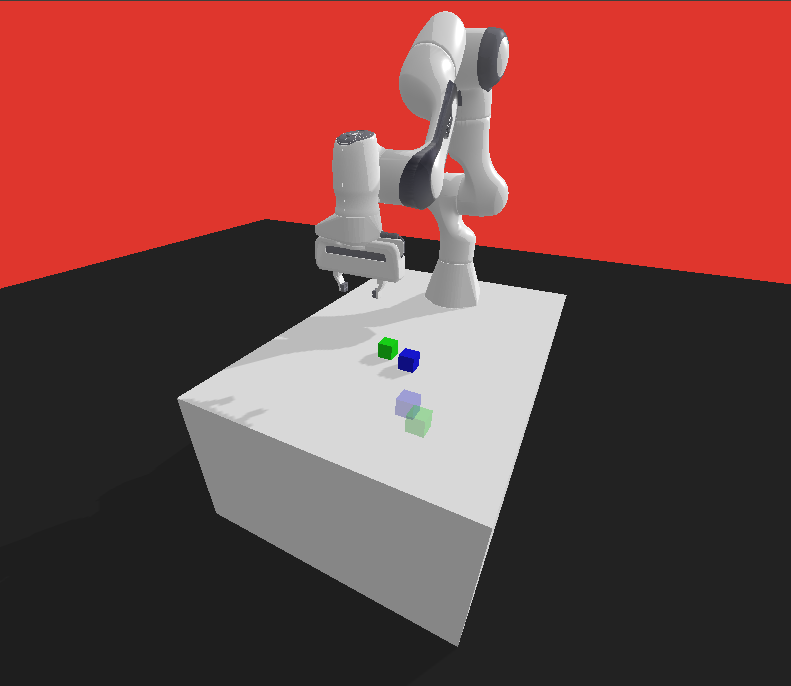}
        \subcaption{Panda arm domain}
        \label{fig:panda-arm}
	\end{minipage}%
	\hfill
	\centering
	\begin{minipage}{.25\textwidth}
		\centering
		\includegraphics[width=\textwidth,height=0.65\textwidth]{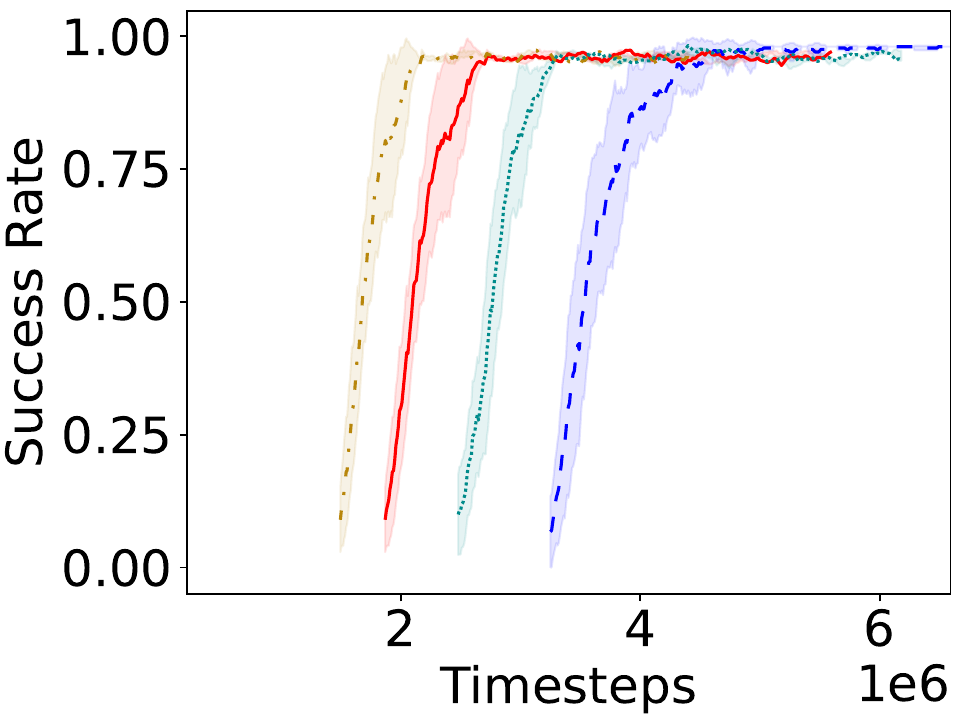}
        \subcaption{Panda arm results}
        \label{fig:lc-panda}
	\end{minipage}%
	\hfill	
 \includegraphics[width=0.8\textwidth,height=0.033\textwidth]{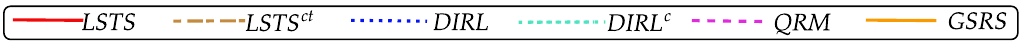}
	\caption{ Learning curves (Averaged over 10 trials) for the two robotic domains.}\label{fig:results-robotics}
\end{figure*}

\subsubsection{LSTS$^{ct}$ (LSTS + Cont. Training) - Gridworld Domain}\label{sec:LSTSct}
In \emph{LSTS}, while learning a policy for $\mathsf{Task}(q,p)$, we reinitialize the environment to a random initial environmental state $s \sim \mathcal{S}_0$ once the agent reaches a state where the propositions ($b_{(q,p)}$) hold true. To answer the question Q2, instead of resetting the environment after reaching such a state where $b_{(q,p)}$ hold true, we let the Teacher agent sample a task (let's say $\mathsf{Task}(p,r)$) from the set $\mathcal{X}[p] \setminus $ {\tt DT}, where $\mathcal{X}$ is the adjacency matrix for the graph, and {\tt DT} is the set of Discarded Tasks, as defined in Algo.~\ref{alg:LSTS}. This helps the agent continue its training by attempting to learn a policy $\pi_{(p,r)}$ for $\mathsf{Task}(p,r)$ while simultaneously learning a separate policy $\pi_{(q,p)}$ for $\mathsf{Task}(q,p)$. If the agent fails to satisfy $\mathsf{Task}(p,r)$, we reset the environment to state $s\sim\mathcal{S}_0$. Otherwise, the agent continues its training until its trajectory satisfies the high-level objective $\phi$.  We call this approach \emph{LSTS}$^{ct}$  (Detailed algo in Appendix A). Results in Table~\ref{table:gridworld} and Fig.~\ref{fig:lc-gridworld} demonstrate that this approach improves sample efficiency by reducing the number of interactions required to learn a successful policy for the gridworld task, with a time-to-threshold metric of $2.7 \times 10^5$ interactions as compared to \emph{LSTS}.

\subsection{LSTS and LSTS$^{ct}$ - Robotic Domains}

To answer (Q3), we test \emph{LSTS} and \emph{LSTS}$^{ct}$ on two simulated robotic environments with high interaction cost. The task in Fig.~\ref{fig:turtlebot} has the following \textsc{SPECTRL} objective:
\begin{multline}
\phi^{navigation}_f := ( \texttt{achieve} (Key_{1}) \:\: \texttt{or} \:\: \texttt{achieve} (Key_{2}) ; \\ \:\: \texttt{achieve} (Goal)) \:\: \texttt{ensuring} (\neg Lava)
\end{multline}
In this task, the agent (a simulated TurtleBot) needs to collect any of the keys (yellow blocks) present in a $[3m,3m]$ continuous environment before reaching the goal position (gray block). At all times, the agent needs to avoid the lava object (red wall) present in the center. 
The move forward (backward) action causes the robot to move forward (backward) by $0.1m$ and the robot rotates by $\pi/8$ radians with each rotate action. The pick-up and drop actions have effects similar to the gridworld domain. The robotic domain is more complex as objects can be placed at continuous locations. The agent receives an ego-centric image view of the environment (top-right corner of Fig.~\ref{fig:turtlebot}), which makes the task partially observable in nature and more complex to get a successful policy. The RL agent is described in Sec.~\ref{sec:results-gridworld}.

The second environment (Fig.~\ref{fig:panda-arm}) consists of a simulated robotic arm pushing two objects to their target locations~\citep{gallouedec2021pandagym} with the \textsc{SPECTRL} formula: 
\begin{align}
\phi^{manipulation}_f :=  \texttt{achieve} (p_{1}) \:\:;\:\: \texttt{achieve} (p_{2})
\end{align}
where $p_1$ and $p_2$ are the atomic propositions for \textit{'push-object-1'}, \textit{`push-object-2'}. The robot has continuous action parameters for moving the arm and a binary gripper action (close/open). An episode begins with the two objects randomly initialized on the table, and the robotic arm has to push these two objects to its final location. The agent receives its current end-effector pose, positions and velocities of the two objects, and the desired goal position for the two objects. For this task, we use the Deep Deterministic Policy Gradient with Hindsight Experience Replay (DDPG-HER)~\citep{andrychowicz2017hindsight} as our RL algorithm. DDPG-HER is implemented using the OpenAI baselines~\citep{baselines}. Both the robotic domains were modeled using PyBullet~\citep{coumans2021}, and the reward structure for both the RL agents was sparse, similar to the one described in Sec.~\ref{sec:results-gridworld}. The learning curves for the TurtleBot domain (Fig.~\ref{fig:lc-turtlebot}) and the Panda arm domain (Fig~\ref{fig:lc-panda}) (averaged over 10 trials) are shown in Fig.~\ref{fig:lc-turtlebot} and Fig.~\ref{fig:lc-panda} respectively. For both domains, \emph{LSTS} outperforms all the baselines in terms of learning speed. \emph{LSTS}$^{ct}$ further speeds-up learning for both the robotic domains. The time-to-threshold between \emph{LSTS} and the best performing baseline (our custom implementation)~\textsc{DIRL}$^c$, is $2 \times 10^6$ for the TurtleBot domain and $5 \times 10^5$ for the Panda arm domain. 

\subsection{LSTS - Search and Rescue task }

To demonstrate how LSTS performs when the plan length becomes deeper, we evaluated LSTS on a complex urban Search and Rescue domain with multi-goal objectives. In this domain, the agent acts in a grid setting where it needs to perform a series of sequential sub-tasks to accomplish the final goal of the task. The agent needs to open a door using a key, then collect a fire extinguisher to extinguish the fire, and then find and rescue stranded survivors.
The order in which these individual sub-goals such as opening the door, rescuing the survivors, and extinguishing the fire are achieved does not matter. A fully-connected graph $\mathcal{G}_{\phi}$ generated using the above mentioned high-level states consists of 24 distinct DAG paths. This is a multi-goal task as the agent needs to find the key to open the door, then extinguish fire and rescue survivors to reach the goal state (details in Appendix F). The results in Table~\ref{table:sr} (averaged over 10 trials) show that \emph{LSTS} reaches a successful policy quicker compared to the LFS, GSRS, QRM and TSCL. The overall number of interactions to learn a set of successful policies for satisfying the high-level goal objective are higher compared to the door key experiment because of the additional complexity of task. We observe that LSTS is able to accommodate the task and learn RL policies that  satisfy the high-level goal objective.

\begin{table}
    		\centering
	\begin{tabular}{>{\centering}p{1.1cm}>{\centering}p{2.6cm}>{\centering}p{2.4cm}p{0.05cm}}
\arrayrulecolor{taupegray}\toprule
\textbf{Approach} & \textbf{$\#$ Interactions} \\ (Mean $\pm$ SD) & \textbf{Success Rate }\\ (Mean $\pm$ SD) \tabularnewline
\midrule
\rowcolor{gainsboro}
\emph{LSTS}  & $(8.61 \pm 0.12)\times 10^6$ &\centering $0.87 \pm 0.04$\tabularnewline
\emph{LFS}  &  $5 \times 10^7$ &\centering $0 \pm 0$\tabularnewline
GSRS  & $5 \times 10^7$ &\centering $0.05 \pm 0.04$ \tabularnewline
QRM  & $5 \times 10^7$ &\centering $0.05 \pm 0.04$ \tabularnewline
TSCL  & $5 \times 10^7$ &\centering $0 \pm 0$ \tabularnewline
\bottomrule
\end{tabular}
		\caption{Table comparing $\#$interactions $\&$ success rate for the Search and Rescue domain.}
		\label{table:sr}
\end{table}

\section{Conclusion}
We proposed \emph{LSTS}, a framework for dynamic task sampling for RL agents using the high-level \textsc{SPECTRL} objective coupled with the Teacher-Student learning strategy. Through experiments, we demonstrated that \emph{LSTS} accelerates learning when the high-level objective is correct and available, converging to a desired success rate quicker as compared to other curriculum learning and automaton-guided RL baselines. Furthermore, \emph{LSTS} does not rely on a human-guided dense reward function. \emph{LSTS}$^{ct}$ further improves sample efficiency by continuing exploration on a new sub-task once a goal state for a sub-task is reached. 

\textbf{Limitations \& Future Work:} In certain cases, the \textsc{SPECTRL} objective can be novel and/or generating the labeling function can be infeasible. Our future plans involve incorporating scenarios where obtaining a precise \textsc{SPECTRL} specification is challenging. If the provided specification does not contain any feasible paths, then the algorithm will not be able to invent new feasible paths. As an extension, we would like to explore biasing away from sub-tasks rather than completely discarding 
them once the target node is reached, so in the limit, optimal policies can be obtained. 


\bibliography{aaai24}

\end{document}